\documentclass[runningheads]{llncs}

\usepackage{eccv}

\usepackage{eccvabbrv}

\usepackage{graphicx}
\usepackage{booktabs}

\usepackage[accsupp]{axessibility}  

\usepackage[pagebackref,breaklinks,colorlinks,citecolor=eccvblue]{hyperref}

\begin{document}

\title{Cycle-Consistency Uncertainty Estimation for Visual Prompting based One-Shot Defect Segmentation} 

\titlerunning{Cycle-Consistency Uncertainty Estimation for Visual Prompting}

\author{Geonuk Kim}


\institute{AI Tech Team, LG Energy Solution\\
\email{geonuk\_kim@korea.ac.kr}}

\maketitle

\begin{abstract}
  Industrial defect detection traditionally relies on supervised learning models trained on fixed datasets of known defect types. While effective within a closed set, these models struggle with new, unseen defects, necessitating frequent re-labeling and re-training. Recent advances in visual prompting offer a solution by allowing models to adaptively infer novel categories based on provided visual cues. However, a prevalent issue in these methods is the over-confdence problem, where models
can mis-classify unknown objects as known objects with high certainty. To addresssing the fundamental concerns about the adaptability, we propose a solution to estimate uncertainty of the visual prompting process by cycle-consistency. We designed to check whether it can accurately restore the original prompt from its predictions. To quantify this, we measure the mean Intersection over Union (mIoU) between the restored
prompt mask and the originally provided prompt mask. Without using complex designs or ensemble methods with multiple networks, our approach achieved a yield rate of 0.9175 in the VISION24 one-shot industrial challenge.
\end{abstract}

\section{Introduction}
\label{sec:intro}

In the realm of industrial defect detection, supervised learning approaches\cite{Jin_2023_CVPR,tabernik2020segmentation} have traditionally dominated due to their ability to leverage labeled datasets to train deep-learning models. These models perform impressively within the confines of a closed set, where the defect categories are pre-defined and remain constant. However, this paradigm faces significant challenges in real-world industrial environments, where new types of defects frequently emerge. The necessity to continuously label and incorporate these novel defects into the training dataset not only imposes a considerable burden but also limits the adaptability of traditional models to unforeseen defect types.

Recent advancements in machine learning have highlighted the potential of visual prompting techniques\cite{li2024visual,jiang2024t}, which offer a promising alternative for handling such dynamic scenarios. Unlike conventional methods that are constrained by fixed labels, visual prompting enables models to dynamically adapt and infer categories not encountered during training. This approach utilizes prompt images—visual cues provided during inference—to guide the model’s interpretation and classification of defects, thereby expanding its capability to handle previously unseen categories. 

However, a prevalent issue in these methods is the overconfdence problem, where models
can mis-classify unknown objects as known objects with high certainty\cite{kim2024improving}. The key expectation from visual prompting is that it enables models to adaptively infer new categories. However, in practice, models often exhibit biases towards previously learned categories, raising fundamental concerns about the adaptability that visual prompting is supposed to offer. To address this issue of bias in visual prompting, we propose a solution where the model outputs a confidence score for the prompting process. We proposed to check whether it can accurately restore the original prompt from its predictions. If the model has inferred the relationship between the prompt and the query image without bias, it should be able to perform accurate reverse inference as well. To quantify this, we measure the mean Intersection over Union (mIoU) between the restored prompt mask and the originally provided prompt mask. This confidence score will help in assessing the reliability of the model’s predictions and mitigate the problem of bias.

By integrating those approach into industrial defect segmentation tasks, we aim to address the inherent limitations of traditional supervised learning methods and visual prompting. The ability of visual prompting to generalize and adapt to new defect types without the need for extensive re-labeling and retraining aligns well with the challenges posed by continuous defect emergence in industrial settings. And further Cycle-consistency based uncertainty estimation enhance the visual prompting reliability.

\section{Baseline Method}
Our baseline is Dinov\cite{li2024visual}, which is a visual prompting method build on top of an encoder-decoder architecture. To effectively formulate visual prompts, they designed prompt encoder to encode reference visual prompts from the reference images and designed shared decoder to decode the final target visual prompts from the target image. They designed an additional prompt classifier to categorize objects within the target images into one of the reference categories. However, the embedding layer trained in this setup undergoes parameter updates that enforce contrastive learning among seen categories within the training set, inherently leading to a bias towards these seen categories.

\section{Proposed method}
Visual prompting models often display biases towards previously learned categories, which raises fundamental concerns regarding the adaptability of visual prompting techniques. To address this issue, we propose to check if it can accurately restore the original prompt from its predictions. 
If the model has inferred the relationship between the prompt and the query image without bias, it should be able to perform accurate reverse inference as well. To quantify this, we measure the mean Intersection over Union (mIoU) between the restored prompt mask and the originally provided prompt mask.

\begin{figure}
  \centering
  \includegraphics[height=6.5cm]{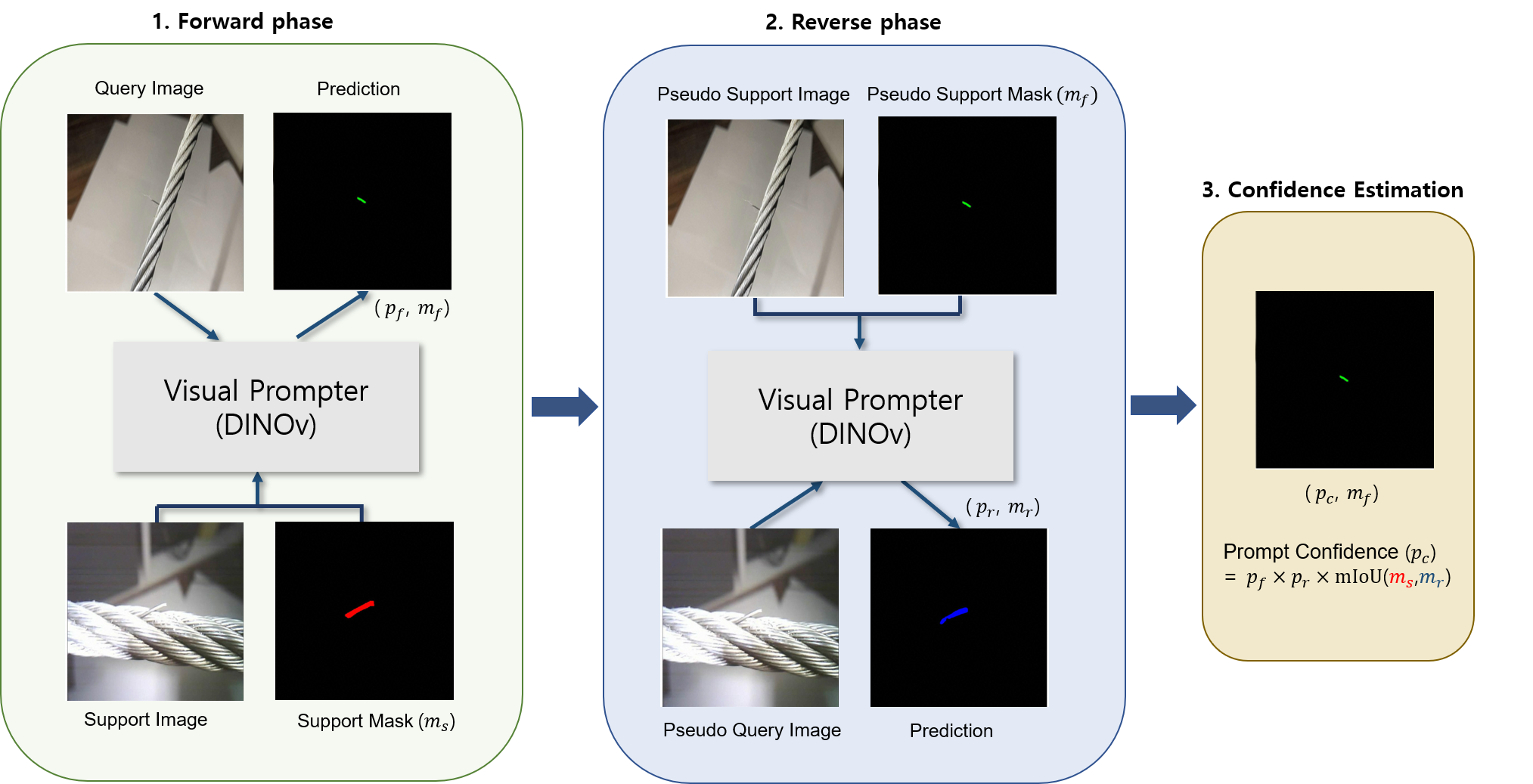}
  \caption{Given a support image with its corresponding prompt mask ms and a query image, the goal of the forward phase is
to identify the regions in the query image that correspond to the prompt. In the
context of segmentation, this process results in the generation of a mask map
\textbf{$m_{f}$} and probability \textbf{$p_{f}$} corresponding to the query image. 
In reverse phase, prompting inference is conducted in reverse.
The query image and the generated mask \textbf{$m_{f}$} are treated as the support image
and support mask, respectively, while the original support image is considered
as the query image. This approach allows for prompting inference to generate a
mask \textbf{$m_{r}$} and \textbf{$p_{r}$} corresponding to the pseudo query image. Subsequently, the mIoU between the original support
mask and the support mask predicted during the reverse phase is computed to
quantify whether the model has made unbiased predictions in both the forward
and reverse phases. 
  }
  \label{fig:example}
\end{figure}

\subsection{Training}
\subsubsection{Image Encoder}
Using strong image feature extractor is a simple way to improve prediction accuracy. We employ modern archtecture Swin-L. We use publicly available pre-trained weights on COCO and ImageNet datasets.
\subsubsection{Data Augmentation}
In data augmentation policy, it is crucial to select methods carefully based on the characteristics of the data. In industrial inspection environments, there is generally variability in illumination but minimal changes in color. To reflect these characteristics, we applied random saturation, random brightness and random contrast with a range of [0.8, 1.2], and performed horizontal flipping of the original images with a probability of 0.5 for training

\subsection{Inference}
Many approaches in almost challenge rely on ensemble methods using multiple modes to achieve high performance on the given test set; however, we did not employ ensemble techniques due to resource constraints to train multiple models. Instead, our goal was to obtain reliable output in a single visual prompting model by estimating the confidence score, which defines how trustworthy the inference of visual prompting model are during the inference stage.
\subsubsection{Forward phase}
As shown in Fig. 1, given a support image with its corresponding prompt mask \textbf{$m_{s}$} and a query image, the goal of the forward phase is to identify the regions in the query image that correspond to the prompt. In the context of segmentation, this process results in the generation of a mask map \textbf{$m_{f}$} and probability \textbf{$p_{f}$} corresponding to the query image.
\subsubsection{Reverse phase}
In reverse phase, prompting inference is conducted in reverse. The query image and the generated mask \textbf{$m_{f}$} are treated as the support image and support mask, respectively, while the original support image is considered as the query image. This approach allows for prompting inference to generate a mask \textbf{$m_{r}$} and \textbf{$p_{r}$} corresponding to the pseudo query image.
\subsubsection{Confidence Estimation}
Subsequently, the mIoU between the original support mask and the support mask predicted during the reverse phase is computed to quantify whether the model has made unbiased predictions in both the forward and reverse phases. Top1 score for each image in both forward and reverse phase then weight the mIoU score. Formally, this is represented as follows:
\begin{equation}
\textit{$p$}_\textit{c} = \textbf{$p_{f}$} \times \textbf{$p_{r}$} \times mIoU(\textbf{$m_{s}$},\textbf{$m_{r}$})
\end{equation}
where \textbf{$p_{f}$} and \textbf{$p_{r}$} represent top-1 score among the matching scores, while \textbf{$m_{f}$} and \textbf{$m_{r}$} denote the corresponding mask map. 
It is notable that existing visual prompting methods exploit \textbf{$p_{f}$} score for inference that increases the number of false positives.

\begin{figure}[h]
  \centering
  \includegraphics[height=8cm]{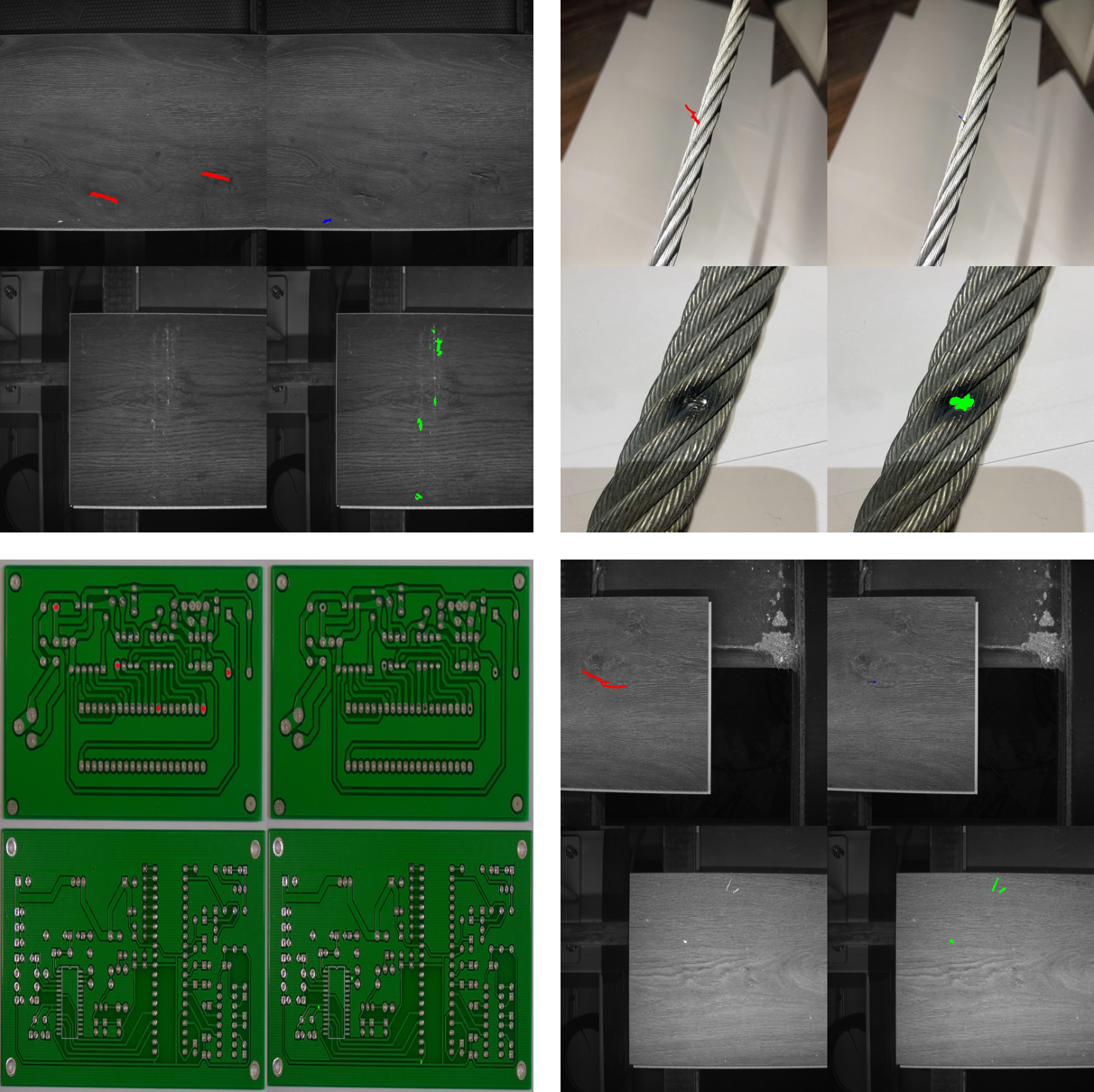}
  \caption{Examples of correct-yield samples corrected by Cycle Consistency-based uncertainty estimation. The red mask in the top left represents the support image and its corresponding ground truth mask map. The bottom left shows the query image. The green mask in the bottom right indicates the query mask inferred through the forward phase, while the blue mask in the top right represents the support mask restored through the reverse phase. In these samples, the support mask was not accurately restored due to model bias, and the $\textit{p}_\textit{c}$ score was lower than the pre-defined threshold, leading the model to convert predicted mask $\textit{m}_\textit{f}$ to null mask.
  In the case of the 'Cable' example, the $\textit{p}_\textit{f}$ value is 0.977, indicating that the model predicted the mask very confidently. However, the mIoU between the restored support mask and the ground truth was measured at 0.048, which is very low.
  }
  \label{fig:example}
\end{figure}

\begin{figure}
  \centering
  \includegraphics[height=8cm]{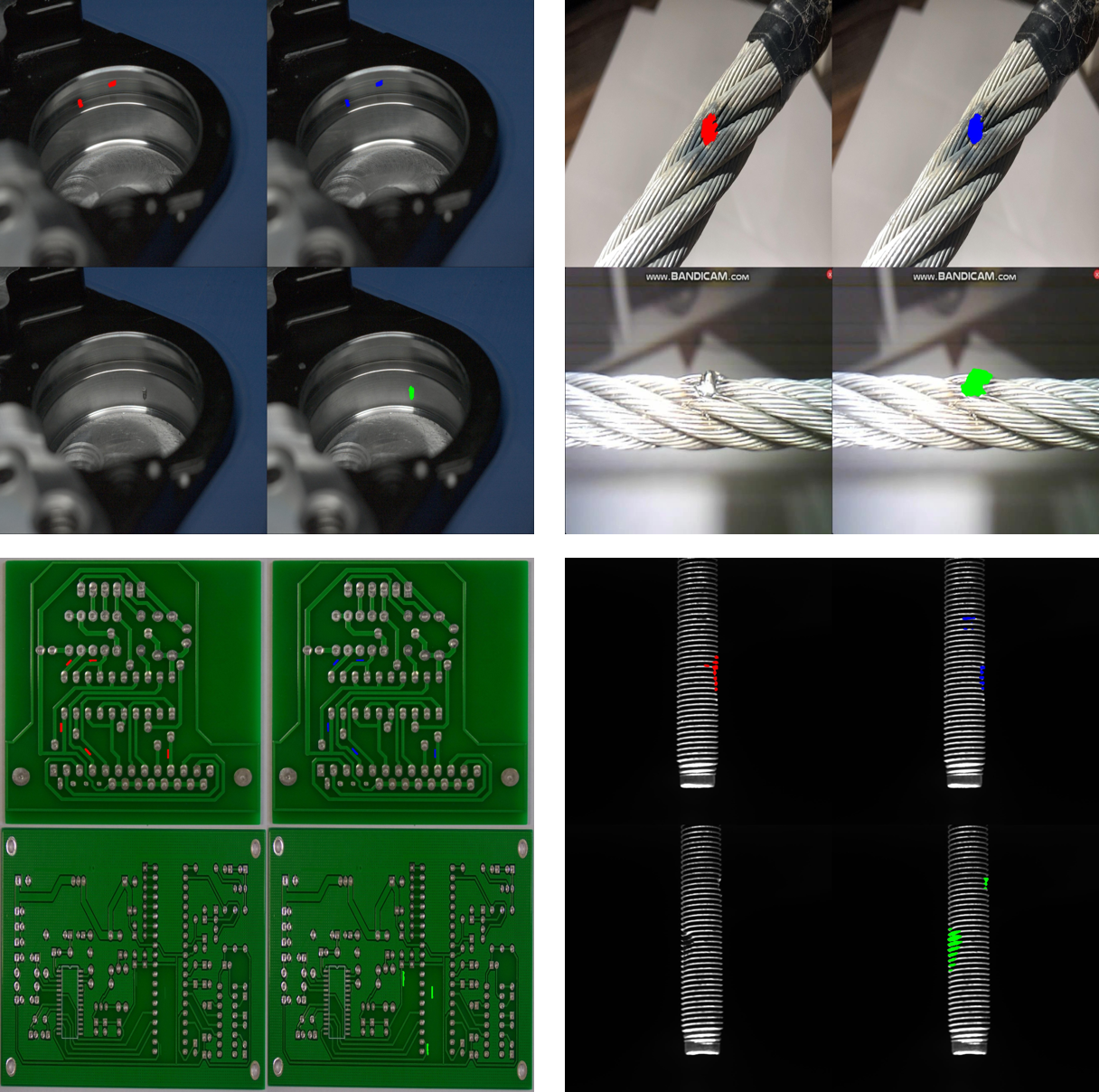}
  \caption{Examples of good-catch samples. The red mask in the top left represents the support image and its corresponding ground truth mask map. The bottom left shows the query image. The green mask in the bottom right indicates the query mask inferred through the forward phase, while the blue mask in the top right represents the support mask restored through the reverse phase. In these samples, the support mask was accurately restored with high mIoU, and the $\textit{p}_\textit{c}$ score was higher than the pre-defined threshold, leading the model to consider predicted $\textit{m}_\textit{f}$ as correct.
  }
  \label{fig:example}
\end{figure}

\section{Experiments}
\subsection{Dataset and Evaluation Metric}
VISION24 one-shot industrial inspection dataset consists of 2024 images for training, 2000 number of support, query pairs for testing. It aims to address the unique data imbalance bottleneck of vision-based industrial inspection by tapping into the potential of reference learning through appropriate visual prompts. Featuring 5 categories of products from diverse domains, this dataset contain 3 groups of defects: known, unknown and unseen defects in the final test set. Rigorously designed evaluation metric evaluates the accuracy of a solution in two key aspects: the positive pair catch rate and the negative pair yield rate. A positive pair is deemed a good catch if the IoU between the predicted mask and the ground truth is greater than or equal to 0.3. For negative pairs, a pair is considered a correct yield if the response rate in the prediction is lower than the pre-defined threshold.
\subsection{Implementation Details}
The training set contains five categories: Cable, Cylinder, PCB, Screw, and Wood, each consisting of one or more defects. For example, the Cable category includes two defects: thunderbolt and torn-apart. We considered categories with different defects, even if they belong to the same main category, as independent classes. Thus, a total of 12 independent classes were defined in the training set. The training images are resized by $800 \times 800 $ and augmented by horizontal flip, random brightness in the range of [0.8, 1.2], random contrast in the range of [0.8, 1.2] and the random saturation in the range of [0.8, 1.2]. The DINOv network is trained with a batch size 64 on 8 GPUs for 20K iterations using AdamW optimizer. All the other settings follow DINOv official implementation.

When the $\textit{p}_\textit{c}$ was greater than 0.18, we trusted and used the mask map predicted by the visual prompting model. When the $\textit{p}_\textit{c}$ was below 0.18, $\textit{m}_\textit{f}$ was converted to a null mask, which led to a tendency for a somewhat lower catch rate while improving yield rate. To improve the catch rate, we used a DINOv model officially pre-trained on COCO and SAM data for this range. By applying cycle-consistency-based uncertainty estimation in the same manner as the DINOv model, we considered the model's prediction as a null mask when was below 0.015, and trusted the predicted mask map when was above 0.015.
  
\subsection{Analysis}
For quantitative evaluation, we assessed the method on the final test set of the challenge and observed the performance shown in Table 1. The proposed method achieved a high yield rate without any specialized network design. This result is quantitatively confirmed by the substantial reduction in false positives due to the Cycle-consistency-based uncertainty estimation. It is anticipated that training multiple models and employing ensemble techniques could further enhance the catch rate by capturing a more diverse feature space.

For qualitative evaluation, we analyzed the results of the forward and reverse phases on the support and query data of the test set. As shown in Fig. 2, in some cases, the support mask was not accurately restored due to model bias, and the $\textit{p}_\textit{c}$ score was lower than the pre-defined threshold, which led the model to correctly convert the predicted mask to a null mask. In the case of the ’Cable’
example, the $\textit{p}_\textit{f}$ value is 0.977, indicating that the model predicted the mask very confidently. However, the mIoU between the restored support mask and the ground truth was measured at 0.048, which is very low.
On the other hand, as shown in Fig. 3, in cases where the predictions were accurate, the model correctly restored the support mask with high mIoU through both the forward and reverse phases. In these samples, the support mask was accurately restored with high mIoU, and the $\textit{p}_\textit{c}$ score was higher than the pre-defined threshold, leading the model to consider the predicted mask. 

\begin{table}[h]
\begin{center}
\centering
\caption{The proposed method achieved a high yield rate without requiring any special network designs or complex ensemble structures. This is quantitatively validated by the significant reduction in false positives due to the Cycle-consistency-based uncertainty estimation. }\label{Tab:zeroshotResult2}

\begin{tabular}{c|c|c|c}
\hline
Model& Catch rate &Yield rate& PES \\ \hline
ours & 	0.77500 &	0.91750 &	0.84625\\ \hline

\end{tabular}
\end{center}
\end{table}

\label{sec:blind}

\section{Conclusion}
Recent advancements in visual prompting allow models to adaptively infer novel categories based on visual cues. However, a common issue with these methods is over-confidence, where models may misclassify unknown objects as known ones with high certainty. To address these concerns about adaptability, we propose a solution that estimates the uncertainty in the visual prompting process through cycle-consistency. Our method involves verifying whether the original prompt can be accurately restored from its predictions. We quantify this by measuring the mean Intersection over Union (mIoU) between the restored prompt mask and the original prompt mask. Experimental analysis demonstrated that false positive masks with high prediction scores could be corrected through cycle-consistency-based uncertainty estimation. Additionally, without employing complex designs or ensemble methods with multiple networks, our approach achieved a yield rate of 0.9175 in the VISION24 one-shot industrial challenge.

%
%
\bibliographystyle{splncs04}
\bibliography{main}
\end{document}